\documentclass[10pt, a4paper]{article}
\usepackage{lrec2000}
\usepackage{epsfig,rotate,alltt,boxedminipage}
\usepackage{url}

\def\smtt#1{{\small\tt #1}}
\def\sburl#1{[{\small\url{#1}}]}
\def\boxfig#1{\fbox{\parbox{0.97\linewidth}{\centerline{\epsfig{#1}}}}}

\newcommand{\queryseq}{-$>$\ }
\newcommand{\querydom}{\^{}\ }
\newcommand{\queryassoc}{$=>$\ }

\newcommand{\ANTE}{\mbox{\it ante}}
\newcommand{\POST}{\mbox{\it post}}
\newtheorem{defn}{Definition}
\newenvironment{sv}{\scriptsize\begin{alltt}}{\end{alltt}\normalsize}
\newenvironment{vb}{\small\begin{alltt}}{\end{alltt}\normalsize}

\title{Towards A Query Language for Annotation Graphs}

{\setlength{\parindent}{0pt}
\name{\noindent
  Steven Bird$^{\ast}$,
  Peter Buneman$^{\dagger}$ and
  Wang-Chiew Tan$^{\dagger}$
}

\address{
$^{\ast}$Linguistic Data Consortium, University of Pennsylvania,
  3615 Market Street, Philadelphia, PA 19104, USA\\
$^{\dagger}$Department of Computer Science,
  University of Pennsylvania, 200 South 33rd Street,
  Philadelphia, PA 19104, USA
}}

\abstract{
The multidimensional, heterogeneous, and temporal nature of
speech databases raises interesting challenges for
representation and query.  Recently, annotation graphs have
been proposed as a general-purpose representational framework
for speech databases.
Typical queries on annotation graphs
require path expressions similar to those used in
semistructured query languages.  However, the underlying model is
rather different from the customary graph models for
semistructured data: the graph is acyclic and unrooted, and both
temporal and inclusion relationships are important.  We develop a
query language and describe optimization techniques for an underlying
relational representation.
}

\begin{document}

\maketitleabstract

\section{Introduction}

In recent years, annotated speech databases have grown
tremendously in size and complexity.  In order to maintain
or access the data, one invariably has to write special
purpose programs.  With the introduction of a general
purpose data model, the annotation graph \cite{BirdLiberman99},
it is possible to abstract away from idiosyncrasies of
physical format.  However, this does not magically solve
the maintenance and access problems.  In this paper, we
contend that some form of query language is essential for
annotation graphs, and we report our research on such
a language.

Query languages for databases have two,
sometimes conflicting, purposes.  First they should express -- as
naturally as possible -- a large number of data extraction and
restructuring tasks.  Second, they should be optimizable.  This means
that they should be based on a few efficiently implemented primitives;
they should also make it easy to discover optimization strategies that may
involve query rewriting, execution planning and indexing.  The
relational algebra and its practical embodiment, SQL, are examples of
such languages, however they are unsuitable for annotation graphs
first because it is difficult (or impossible -- depending on the
version of SQL) to express many practical queries, and second because
the optimizations that are necessary for annotation graph queries are
not in the repertoire of standard relational query optimizations.

The recent development of query languages for semistructured data
\cite{Buneman96,Quass95,xml-ql}
offer more natural forms of expression for
annotation graphs.  In particular, these languages support {\em
regular path patterns} -- regular expressions on the labels in the
graph -- to control the matching of variables in the query to vertices
or edges in the graph.  While regular path patterns are useful, the
usual model of semistructured data, that of a labeled tree, is not
appropriate for annotation graphs.  In particular, it fails to capture
the quasi-linear structure of these graphs, which is essential in
query optimization.

After reviewing some existing languages for linguistic annotations,
we present the annotation graph model, its relational representation,
and some relational queries on annotation graphs.  Then
we develop a new query language for annotation graphs that allows complex
pattern matching.  It is loosely based on semistructured query
languages, but the syntax simplifies the problem of finding regions of
the data that bound the search.  Finally, we describe an
optimization method that exploits the quasi-linear
structure of annotation graphs.

\section{Query Languages for Annotated Speech}

If linguistic annotations could be modeled as simple hierarchies,
then existing query languages for structured text would apply
\cite{Clarke95,Sacks-Davis97}.  However, it is possible to have
independent annotations of the same signal (speech or text) which
chunk the data differently.  As a simple example, the division
of a text into sentences is usually incommensurable with its
division into lines.  Such structures cannot be represented
using nested, balanced tags.

The fundamental problem faced by any general purpose query language
for linguistic annotations is the navigation of these multiple
intersecting hierarchies.  In this section we consider two
query languages which address this issue.

\subsection{The Emu query language}

\begin{figure}[ht]
\boxfig{figure=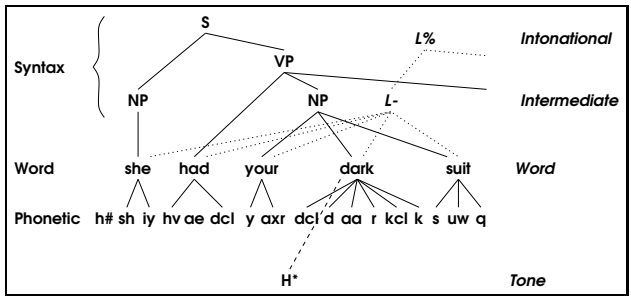,width=\linewidth}
\caption{Intersecting Hierarchies in Emu}
\label{fig:emu}
\end{figure}

The Emu speech database system \sburl{www.shlrc.mq.edu.au/emu}
\cite{CassidyHarrington96,CassidyHarrington99} provides tools for creation
and analysis of data from annotated speech databases.  Emu
annotations are arranged into levels (e.g. phoneme, syllable,
word), and levels are organized into hierarchies.  Emu supports
multiple independent hierarchies, such that any specific level
may participate in more than one orthogonal structure.  An example
is shown in Figure~\ref{fig:emu} \cite{CassidyBird00}.

A database of such annotations can be searched using the
Emu query language.  The language has primitives for
sequence, hierarchy and ``association'', as illustrated
below.

\begin{list}{}{\setlength{\itemindent}{-\leftmargin}\setlength{\rightmargin}{0pt}\setlength{\itemsep}{0pt}}

\item \smtt{[Phonetic=a|e|i|o|u]} --
matches a disjunction of items on the phonetic level

\item \smtt{[Phonetic=vowel \queryseq Phonetic=stop]} --
matches a sequence of \smtt{vowel} followed immediately by \smtt{stop}.

\item \smtt{[Word!=dark \querydom Phoneme=vowel]}
matches an word not labelled \smtt{dark} immediately dominating \smtt{vowel}.

\item \smtt{[Word!=x \queryassoc Tone=H*]}     
  Find any word associated\footnote{
    This ``association'' can have either a temporal interpretation as
    overlap \cite{BirdKlein90} and an atemporal interpretation
    as some essentially arbitrary binary relation; both
    interpretations are encompassed by our model.
  }
  with a \smtt{H*} tone
\end{list}

Note that the language lacks a wildcard, and \smtt{Word!=x}
serves this purpose in the absence of any actual word \smtt{x}.

More complex queries are built up using nesting.  There is
no (non-atomic) disjunction or negation in the language.
An example of a nested query follows;
here, the query finds any syllable dominating a stop that precedes
a vowel which is associated to a high tone.

\begin{vb}
 [Syllable=S ^ 
     [Phonetic=stop -> 
         [Phonetic=vowel => Tone=H*]]]
\end{vb}

Cassidy has shown how expressions of
this query language can be translated into a first-order
query language, in this case, SQL \cite{Cassidy99}.

In the Emu query language, the dominance relation is symmetric.
(A separate type hierarchy is used to order the levels.)
This property makes it possible to navigate a path through
multiple hierarchies without using variables.  For example,
The following expression finds an \smtt{NP} which dominates
a word \smtt{dark} that is dominated by an intermediate phrase
that bears an \smtt{L-} tone.\footnote{We are grateful to Steve
  Cassidy for providing this example.}

\begin{vb}
[ syntax=NP ^ [ word=dark ^ intermediate=L- ]]
\end{vb}

These expressions correspond to the ``where'' clause of
a conventional query language.  The Emu query language lacks
an explicit ``select'' clause.  Rather the selected material is
the left-most element of the where clause, by default, or else
the single element distinguished with a hash prefix.  The
query result is a column of these elements, and this is typically
processed with an external statistics package.

\subsection{The MATE query language}

\begin{figure}[t]
\boxfig{figure=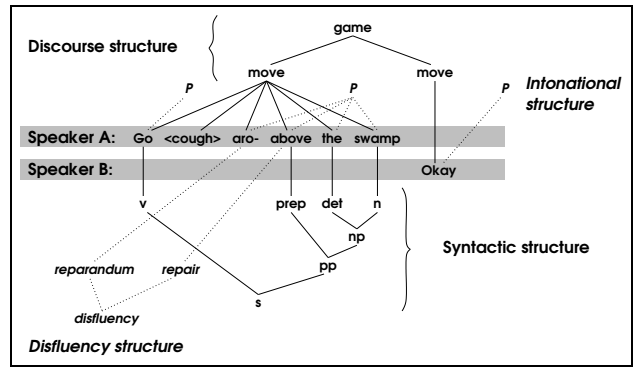,width=\linewidth}
\caption{Intersecting Hierarchies in MATE}
\label{fig:mate}
\end{figure}

The MATE project is developing standards and tools for annotating
spoken dialogue corpora \sburl{mate.nis.sdu.dk}.  Like Emu, MATE
supports intersecting hierarchies; Figure~\ref{fig:mate} illustrates
four hierarchies built over the same dialogue transcript
\cite{Carletta99}.  These hierarchies happen to intersect at their
fringe, however this need not be the case.

MATE uses XML to represent these structures.  Each node in
Figure~\ref{fig:mate} corresponds to an XML element, and the
node labels correspond to an attribute of the element or its
content.  For example, \smtt{swamp} could be represented as
\smtt{<word id="A6" num="sing">swamp</word>}, and \smtt{np}
could be represented as \smtt{<phrase type="np"/>}.
In the query language \cite{Mengel99}, we can pick out these elements with
the following expressions:

\begin{sv}
($w word); $w.orth ~ "swamp"
($p phrase); $p.type ~ "np"
\end{sv}

Hierarchical
relationships, like the one between \smtt{game} and \smtt{move}
or between \smtt{move} and \smtt{swamp}, are represented
using nesting of XML elements or by hyperlinks.
The query language has a transitive dominance relation \smtt{\^{}}
which navigates down through nested structures and hyperlinks.
For example, we can find noun phrases dominating the word
``swamp'' with the expression:

\begin{sv}
($p phrase) ($w word);
  ($p.type ~ "np")
  && ($w.orth ~ "swamp")
  && ($p ^ $w)
\end{sv}

Each element spans an extent of textual material, and the
query language supports a variety of temporal comparisons
on these extents, reminiscent of Allen's temporal relations
\cite{Allen83}.  So long
as two hierarchies intersect at their terminals (and not at
non-terminals) then their non-terminals will be comparable
using these temporal expressions.  However, the language
directly supports queries on intersecting hierarchies.
For example, we can find a word which is simultaneously a repair and a
preposition, where \smtt{1\^{}} is the immediate dominance
relation:\footnote{We thank David McKelvie for furnishing this example.}

\begin{sv}
($w word) ($ph phrase) ($r repair) ($d disfluency); 
  ($r 1^ $w) && ($ph 1^ $w)
   && ($ph type ~ "prep") && ($d 1^ $r)
\end{sv}

Unlike the Emu query language, the formal and computational
properties of the MATE query language, vis-\`a-vis relational and
semistructured query languages, are unexplored.

This concludes our brief survey of query languages for annotated
speech.  Other query languages exist; these two were chosen because
of their interesting approach to the problem of intersecting
hierarchies.

\section{Annotation Graphs}
\label{ag-defs}

Annotation Graphs were presented by Bird and Liberman
as follows.  Here we consider just the so-called ``anchored'' variety.

\begin{defn}
An \textbf{anchored annotation graph} $G$ over a label set $L$ and
timelines $\left<T_i, \leq_i\right>$ is a 3-tuple
$\left< N, A, \tau \right>$ consisting of a node set $N$,
a collection of arcs $A$ labeled with elements of $L$,
and a time function $\tau: N \rightharpoonup \bigcup T_i$,
which satisfies the following conditions:

\begin{enumerate}\setlength{\itemsep}{0pt}

\item $\left< N, A \right>$ is a labeled acyclic digraph
  containing no nodes of degree zero;

\item for any path from node $n_1$ to $n_2$ in $A$,
  if $\tau(n_1)$ and $\tau(n_2)$ are defined, then
  there is a timeline $i$ such that
  $\tau(n_1) \leq_i \tau(n_2)$;

\item If any node $n$ does not have both incoming and
  outgoing arcs, then $\tau: n \mapsto t$ for some time $t$.

\end{enumerate}
\end{defn}

Note that annotation graphs may be disconnected or empty, and that they must
not have orphan nodes.  It follows from the above definition
that every node has two bounding times, and we will make
use of this property later.  It also follows from the definition
that timelines partition the node set.

\begin{figure}[b]
\begin{boxedminipage}[t]{\linewidth}
\begin{sv}
train/dr1/fjsp0/sa1.wrd:   train/dr1/fjsp0/sa1.phn:
2360 5200 she              0 2360 h#
5200 9680 had              2360 3720 sh
9680 11077 your            3720 5200 iy
11077 16626 dark           5200 6160 hv
16626 22179 suit           6160 8720 ae
22179 24400 in             8720 9680 dcl
24400 30161 greasy         9680 10173 y
30161 36150 wash           10173 11077 axr
36720 41839 water          11077 12019 dcl
41839 44680 all            12019 12257 d
44680 49066 year           ...
\end{sv}
\centerline{\epsfig{figure=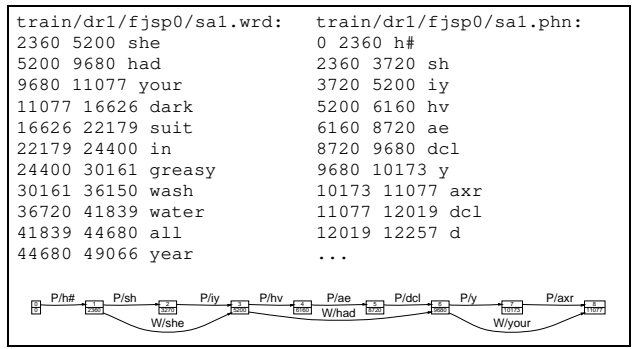,width=0.95\linewidth}}
\end{boxedminipage}
\caption{TIMIT Annotation Data and Graph Structure}\label{fig:timit}
\vspace*{2ex}
\end{figure}

The formalism can be illustrated with an application to a simple
speech database, the TIMIT corpus of read speech \cite{TIMIT86}.
This database contains recordings of 630 speakers of 8 major
dialects of American English, each reading 10 phonetically rich sentences
\sburl{www.ldc.upenn.edu/Catalog/LDC93S1.html}.
Figure~\ref{fig:timit} shows part of the annotation of one of the
sentences.  The file on the left contains word transcription, and
the file on the right contains phonetic transcription.  Part of
the corresponding annotation graph is shown underneath.
Each node displays the node identifier
and the time offset (in 16kHz sample numbers).  The arcs are decorated
with type and label information.  The type \smtt{W} is for words and
the type \smtt{P} is for phonetic transcriptions.

\begin{figure*}[p]
\begin{vb}
962.68 970.21 A: He was changing projects every couple of weeks and he
  said he couldn't keep on top of it. He couldn't learn the whole new area  
968.71 969.00 B: %mm.  
970.35 971.94 A: that fast each time.  
971.23 971.42 B: %mm.    
972.46 979.47 A: %um, and he says he went in and had some tests, and he
  was diagnosed as having attention deficit disorder. Which  
980.18 989.56 A: you know, given how he's how far he's gotten, you know,
  he got his degree at &Tufts and all, I found that surprising that for
  the first time as an adult they're diagnosing this. %um  
989.42 991.86 B: %mm. I wonder about it. But anyway.  
991.75 994.65 A: yeah, but that's what he said. And %um  
994.19 994.46 B: yeah.  
995.21 996.59 A: He %um  
996.51 997.61 B: Whatever's helpful.  
997.40 1002.55 A: Right. So he found this new job as a financial
  consultant and seems to be happy with that.  
1003.14 1003.45 B: Good.  
\end{vb}
\vspace*{2ex}

\centerline{\epsfig{figure=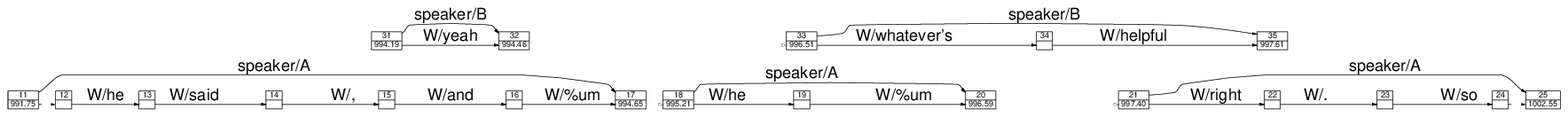,width=\linewidth}}
\vspace*{2ex}\hrule
\caption{CALLHOME Telephone Speech Data and Graph Structure}\label{fig:callhome}
\vspace*{4ex}

{\small
\begin{minipage}{\textwidth}
{\it Arc}
\begin{tabular}[t]{rrrc}
 $\underline{A}$ &
 $X$  &
 $Y$  &
 $T$ \\
\hline

1 & 0 & 1 & P \\
2 & 1 & 2 & P \\
3 & 2 & 3 & P \\
4 & 3 & 4 & P \\
5 & 4 & 5 & P \\
6 & 5 & 6 & P \\
7 & 6 & 7 & P \\
8 & 7 & 8 & P \\
9 & 8 & 9 & P \\
10 & 9 & 10 & P \\
11 & 10 & 11 & P \\
12 & 11 & 12 & P \\
13 & 12 & 13 & P \\
14 & 13 & 14 & P \\
15 & 14 & 15 & P \\
16 & 15 & 16 & P \\
17 & 16 & 17 & P \\
\end{tabular}\hspace{2ex}
\begin{tabular}[t]{rrrc}
 $\underline{A}$ &
 $X$  &
 $Y$  &
 $T$ \\
\hline
19 & 3 & 6 & W \\
20 & 6 & 8 & W \\
21 & 8 & 14 & W \\
22 & 14 & 17 & W \\
23 & 1 & 18 & S \\
24 &3 & 18 & S \\
25 &1 & 3 & S \\
26 &3 & 6 & S \\
27 &6 & 17 & S \\
28 &1 & 17 & Imt \\
29 &1 & 18 & Itl \\
30 &1 & 19 & T \\
31 &19 & 20 & T
\end{tabular}\hfil
{\it Time}
\begin{tabular}[t]{rr}
 $\underline{N}$ &
 $T$\\
\hline
0 & 0    \\
1 & 2360 \\
2 & 3270 \\
3 & 5200 \\
4 & 6160 \\
5 & 8720 \\
6 & 9680 \\
7 & 10173\\
8 & 11077\\
9 & 12019\\
10 & 12257\\
11 & 14120\\
12 & 15240\\
13 & 16200\\
14 & 16626\\
15 & 18480\\
16 & 20685\\
17 & 22179
\end{tabular}\hfil
{\it Label}
\begin{tabular}[t]{rc}
 $\underline{A}$ &
 $L$  \\
\hline
1 & h\# \\
2 & sh  \\
3 & iy  \\
4 & hv  \\
5 & ae  \\
6 & dcl \\
7 & y   \\
8 & axr \\
9 & dcl \\
10 & d \\
11 & aa \\
12 & r \\
13 & kcl \\
14 & k \\
15 & s \\
16 & uw
\end{tabular}\hspace{2ex}
\begin{tabular}[t]{rc}
 $\underline{A}$ &
 $L$  \\
\hline
17 & q \\
18 & she \\
19 & had \\
20 & your \\
21 & dark \\
22 & suit \\
23 & S \\
24 & VP \\
25 & NP \\
26 & V \\
27 & NP \\
28 & L- \\
29 & L\% \\
30 & 0 \\
31 & H* 
\end{tabular}
\end{minipage}    
}
\vspace*{2ex}\hrule
\caption{The Arc, Time and Label Relations}
\label{fig:graph-table}
\vspace*{4ex}

\centerline{\epsfig{figure=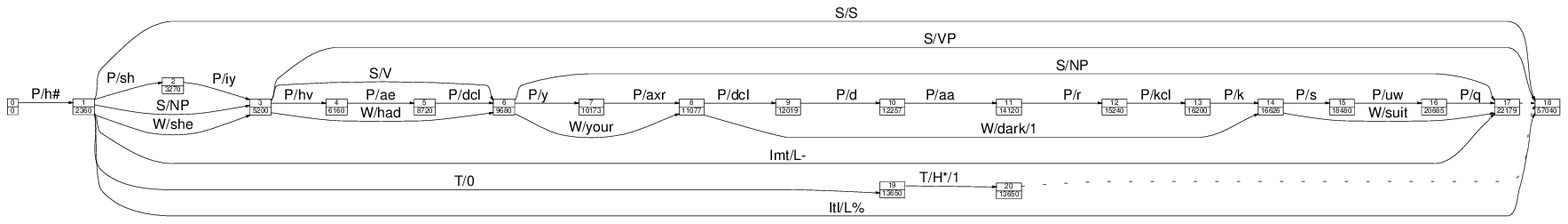,width=\linewidth}}
\vspace*{2ex}\hrule
\caption{Annotation Graph for Extended TIMIT Example}
\label{fig:ag-timit}
\end{figure*}

Observe that all the nodes in Figure~\ref{fig:timit} have time
values.  This need not be the case.  For example, in
the CALLHOME telephone speech corpus
\sburl{www.ldc.upenn.edu/Catalog/LDC96S46.html}, times are
only available for speaker-turn boundaries (see Figure~\ref{fig:callhome}).

Annotations expressed in the annotation graph data
model can be trivially recast as a set of relational tables
\cite{CassidyBird00}, just as can be done for semistructured
data \cite{Florescu99}.  We employ three relations: {\it arc},
{\it time} and {\it label}.
The arc relation is a four-tuple
containing an arc id, a source node id, a target node id, and a type.
The time relation maps (some of) the node ids to times.  The label
relation maps the arc ids to labels.

Figure~\ref{fig:graph-table} gives an instance of this schema
for the TIMIT data of Figure~\ref{fig:timit} (enriched with the
information shown in Figure~\ref{fig:emu}.  The names of key
attributes are underlined.  Figure~\ref{fig:ag-timit} shows
the graph representation for this data.  Note that intersecting
hierarchies find a natural expression in this model.

\section{Some Example Queries}

Interesting cases for query are those that involve more than
one of these primitives.  Here are some simple queries to
select subsets of the data.

\begin{enumerate}
\item
  Find word arcs whose phonetic transcription contains a 'd' and
  ends with a 'k'.

\item
  Find phonetic arcs which immediately precede a vowel that
  overlaps a high tone.

\item
  Find words dominating a vowel which overlaps a high tone.

\end{enumerate}

\begin{figure}[b]
\boxfig{figure=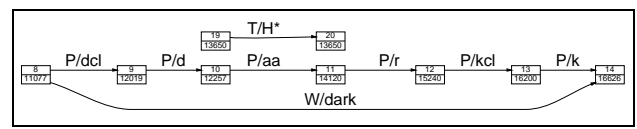,width=\linewidth}
\caption{An Annotation Graph Fragment}
\label{fig:ag-timit2}
\end{figure}

These queries can be interpreted against the fragment
shown in Figure~\ref{fig:ag-timit2}.

Such queries have a first-order interpretation in graphlog
\cite{Consens90}.  We employ a datalog syntax and the
relations in Figure~\ref{fig:graph-table}.  We begin by
defining some auxiliary relations.

First we define a path relation that is sensitive to arc types.
Two nodes \smtt{X} and \smtt{Y}
are connected by a path of type \smtt{T} if there is a sequence
of zero or more arcs, all of type \smtt{T}, beginning at \smtt{X}
and ending at \smtt{Y}.

\begin{sv}
path(X,X,T) :- arc(_,X,_,T)
path(X,X,T) :- arc(_,_,X,T)
path(X,Y,T) :- arc(_,X,Z,T), path(Z,Y,T)
\end{sv}

An arc \smtt{A} ``structurally includes'' an arc \smtt{B} if
there is a path from the start node of \smtt{A} to the start node
of \smtt{B}, and a path from the end node of \smtt{B} to the
end node of \smtt{A}.

\begin{sv}
s_incl(A, B) :- arc(A, X1, Y1, _),
                arc(B, X2, Y2, _),
                path(X1, X2, _),
                path(Y2, Y1, _)
\end{sv}

Finally, an arc \smtt{A} ``temporally overlaps'' an arc \smtt{B}
if the start node of \smtt{A} precedes the end node of \smtt{B},
and the start node of \smtt{B} precedes the end node of \smtt{A}.
(See section~\ref{optimization} for details of the precedence
relation.)

\begin{sv}
ovlp(A, B) :- arc(A, X1, Y1, _), arc(B, X2, Y2, _),
              time(X1, X1t), time(X2, X2t),
              time(Y1, Y1t), time(Y2, Y2t),
              X1t \(\leq\) Y2t, Y1t \(\leq\) X2t
\end{sv}

Now we can provide translations for the three queries
listed above.

\begin{enumerate}
\item
Find word arcs whose phonetic transcription contains a 'd' and
ends with a 'k'.
We assume a relation path/3 which is the transitive closure of arc/4.

\begin{sv}
ans(A) :- arc(A, X, Y, word),
          path(X, X1, phonetic),
          arc(A1, X1, X2, phonetic), label(A1, d),
          path(X2, X3, p),
          arc(A2, X3, Y, phonetic), label(A2, k)
\end{sv}

\item
Find phonetic arcs which immediately precede a vowel that
overlaps a high tone:

\begin{sv}
ans(A) :- arc(A, X, Y, phonetic),
          arc(A1, Y, Y1, phonetic), label(A1, [aeiou]),
          arc(A2, Z, Z1, tone), label(A2, h*)
          ovlp(A1, A2)
\end{sv}

\item
Find words dominating a vowel which overlaps a high tone:

\begin{sv}
ans(A) :- arc(A, _, _, word),
          arc(A1, _, _, phonetic), label(A1, [aeiou]),
          arc(A2, _, _, tone), label(A2, h*),
          s_incl(A, A1), ovlp(A1, A2)
\end{sv}
\end{enumerate}

While it is possible to give queries a first-order interpretation,
the language is quite cumbersome, and we seek a more natural way
to describe annotation graphs.

\section{Query Syntax}

In this section we introduce a query syntax which provides first an
abbreviated notation for the queries expressed previously in datalog.  
Most importantly, the syntax allows us to recognize certain crucial
optimizations.

\subsection{Queries over arc data}

The fundamental unit on which  our query language is built is the arc.
 We form the join of the arc and label relations from
Figure~\ref{fig:graph-table} and adopt names for our attributes.
A query that finds the arc identifiers, types and labels of
all edges in timeline \smtt{tl1} is shown below:

\begin{sv}
select ans(E,T,L)
where [id: E, type: T, label: L] <- tl1
\end{sv}

We follow the datalog convention of using uppercase symbols for
variables and lowercase symbols for constants.
The notation \smtt{[id: E, type: T, label: L]} is used for arcs and
describes  a  {\em arc pattern}: it is matched against the
arcs in the timeline \smtt{tl1} and binds the variables \smtt{E,T,L}
for each match to the arc data in the timeline.  For each such match
it constructs a tuple \smtt{ans(E,T,L)} in the output.  Arc patterns
may contain constants, e.g. \smtt{[id: E, type: word, label: L]} and
there is no constraint on their width.  In this sense they are
"ragged" or "semistructured" tuples.

\begin{sv}
[id: E, start: X, end: Y, type: T, name: N,
  xref: X, lex-id: L, annotator: SB]
\end{sv}

Since attributes are distinguished by name rather than position, it
is safe (and often convenient) to omit them when we do not need
to constrain their value, or bind a variable.

To query over a collection of timelines \smtt{timit} we use cascaded
bindings:

\begin{sv}
select ans(E,L)
where  TL <- timit
       [id: E, start: X, end: Y, label: L, type: word] <-TL
       time(Y) - time(X) < 8000
\end{sv}
This selects the edge identifiers and labels (the names of the words)
from all words in the timit corpus of a suitably short duration.

The form of this query follows a standard syntax for semistructured
query languages (see \cite{Abiteboul00}).   We shall concentrate here on
the development of a syntax for patterns that specify paths and assume
a standard syntax, e.g. \smtt{select ans(E,L)}, for returning results
of the query.

\subsection{Path patterns}

Each arc has a start and end node.   We can specify two adjacent arcs
by requiring the start node of one arc to be the end node of another

\begin{sv}
[id: E1, start: X, end: Y, type: T1, label: L1] <- db
[id: E2, start: Y, end: Z, type: T2, label: L2] <- db
\end{sv}

In this fashion we can specify any sequence of arcs.   However we shall
use an abbreviated syntax \smtt{[ ... ].[ ... ]} to specify the
concatenation of edges, that is, the dot is an associative pattern
concatenation operation. Thus the previous pair of patterns binds the
same variables as the following single pattern

\begin{sv}
[id: E1, start: X, end: Y, type: T1, label: L1] .
[id: E2, end: Z, type: T2, label: L2] <-  db
\end{sv}

Within edge patterns we also allow arbitrary predicates.  For
example:
\smtt{[type: T, T = word or T = ph]},  \smtt{[start: X, stop: Y,
time(Y) - time(X) > 200]}.  Predicates may also use attribute names as
values. For example,  \smtt{[type: word]}, \smtt{[type: X, X = word]},
\smtt{[type=word]} are equivalent.

A sequence of arcs (phonemes, syllables, phrases, etc) is represented
in our model using a concatenated sequence of arc patterns.  To
specify path patterns of arbitrary length 
we also allow arbitrary regular expressions on arcs.  
An arbitrary path of word arcs is represented by 
\smtt{[type = word]*} and an arbitrary path of word or phoneme arcs
by \smtt{([type = word]|[type = phoneme])*}.  Care must be taken
in interpreting variables inside a Kleene * or a union.  The rule is
that such variables must be bound elsewhere in the program. We cannot 
bind variables inside a union or Kleene *.  Thus \smtt{[type: T]*} is
illegal. 

Suppose we have a path pattern \smtt{[type: word]*} and we want to
refer to the first node on the path.   The pattern 
\smtt{[start: X, type: word]*} is illegal.  (Even if it were legal
this pattern could only only match paths of length 0 or 1.)
To allow the binding of nodes outside of an edge pattern we take
single variables in the sequence to denote nodes.  For example,
\smtt{X.[type: T].Y} is equivalent to \smtt{[start: X, type: T, end: Y]}.
Moreover, \smtt{X.[type: word]*.Y} binds \smtt{X} to the first node
and \smtt{Y} to the last node on a path of \smtt{word} arcs.
Now consider the following example:

\begin{sv}
X.[type = parse, label = sentence].Y <- db           (a)
X.[type = word]*.[type = word, label = opera]
 .[type = word]*.Y <- db
\end{sv}
This matches the start and end node of any sequence that contains the word
\smtt{opera}. Another possibility is shown below.

\begin{sv}
X.[type = parse, label = sentence].Y <- db           (b)
X'.[type = word, label = opera].Y' <- db
time(X) <= time(X') and time(Y') <= time(Y)
\end{sv}
However, (a) and (b) are {\em not} equivalent queries.

\begin{figure}[t]
\boxfig{figure=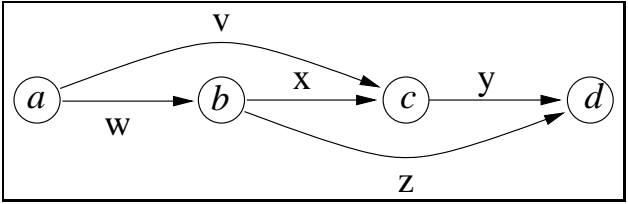,width=\linewidth}
\caption{An Annotation Graph whose Description Requires Variables}
\label{fig:figure8}
\end{figure}

One might think, from example (a) above, that one could dispense with
node variables by having a parallel composition operator.
It turns out that there are many situations where this is
impossible.  The simplest instance is shown in
Figure~\ref{fig:figure8}.

The annotation graph in Figure~\ref{fig:figure8} cannot
be uniquely described using parallel and serial composition.
Instead, we need a set of expressions as follows:

\begin{sv}
A.[label: W].B.[label: X].C.[label: Y].D <- db
A.[label: V].C <- db
B.[label: Z].D <- db
\end{sv}

\subsection{Arbitrary predicates on arcs}

The bracket notation for describing arcs can also
enclose arbitrary predicates.  Predicates expressing the (temporal)
overlap or inclusion of edges are particularly useful. Example (b)
above may be expressed as.

\begin{sv}
[id: E, type = parse, label = sentence] <- db
[type = word, label = opera, subinterval(E)] <- db
\end{sv}

Note that \smtt{subinterval(E)} can be thought of as a ``method'' of
the edge, that is called when the pattern is matched.

\subsection{Abbreviations}  

The preceding syntax is quite general; it has little to do with the
specific conventions of linguistic data. Paths typically, though not
always, follow the same type.  Labels are also special.  We propose
the following syntactic sugar.  (The proposal is tentative, all sorts
of variations are possible).

Given a database of arcs {\em db}, the notation {\tt db/t} restricts the
database to those arcs of type $t$.  Also the notation \smtt{:L} is an
abbreviation for \smtt{label: l}.  For
example, \smtt{X.[:L].Y <- db/word} is shorthand for
\smtt{X.[label: L, type: word].Y <- db}.  Using this, example (a)
becomes:

\begin{sv}
X.[:sentence].Y <- db/parse                         (a')
X.[]*.[:opera].[]*.Y <- db/word
\end{sv}

\subsection{Horizontal path expressions}

Find words with {\tt c.*t.*} (our first query)

\begin{sv}
X.[].Y <- db/word
X.[:c].[]*.[:t].[]*.Y <- db/ph
\end{sv}

Here's a harder case, with a variable inside the
scope of a Kleene star.  The predicate \smtt{ovlp(E)} is an ``overlap''
predicate. 

\begin{sv}
X.[].Y <- db/word
[id: E] <- db/background
X.[:c].[ovlp(E)]*.[:t].[]*.Y <- DB/ph
\end{sv}

In this section we have paid little attention to the output of a
query.  From the introductory examples, it should be clear that it is
straightforward to construct a set of tuples in the same sense that
datalog constructs a set of tuples.  It is also possible to extend the
syntax to express the construction and augmentation of annotation
graphs.  The details will be described elsewhere

\section{Optimization: exploiting quasi-linearity}
\label{optimization}

\begin{figure}[t]
\boxfig{figure=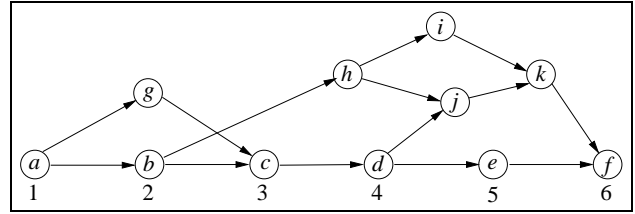,width=\linewidth}
\caption{A precedence graph}\label{fig:pgraph}
\end{figure}

In the previous sections we developed a query language for annotation
graph data
and showed how an analysis of that language might help - in many
practical cases -- to lead to tractable implementations. Here we show
how we can exploit the ``almost sequential'' notion of annotation graph data to
support these implementations.  In particular, we will show how to use
the underlying temporal order to select a small fragment of the input
data that will fit into main memory, bypassing many of the
database optimization issues.

\begin{figure}[t]
\begin{boxedminipage}[t]{\linewidth}
{\small
\begin{center}
\(
\mbox{\it time}\ \ \
\begin{array}[t]{llll}
\mbox{node} & \mbox{timeline}  & \mbox{ante} & \mbox{post}\\ \hline
a & T_1 & 1 & 1\\
b & T_1 & 2 & 2\\
c & T_1 & 3 & 3\\
d & T_1 & 4 & 4\\
e & T_1 & 5 & 5\\
f & T_1 & 6 & 6\\
g & T_1 & 1 & 3\\
h & T_1 & 2 & 6\\
i & T_1 & 2 & 6\\
j & T_1 & 4 & 6\\
k & T_1 & 4 & 6
\end{array}
\)
\hfil
\(
\mbox{\it TA'}\ \ \ \begin{array}[t]{ll}
\mbox{source}&\mbox{target}\\ \hline
h & i\\
h & j\\
j & k\\
i & k\\
h & k\\
\end{array}
\)
\end{center}
}
\end{boxedminipage}
\caption{The Time and TA Relations}
\label{fig:ta}
\end{figure}

Consider the example in Figure~\ref{fig:pgraph}.   It shows a collection
of nodes, where nodes a-f are timed and the rest are untimed.  All the
untimed nodes are linked by
arcs to other nodes.  In order to extract those portions of the
database that are needed to answer a query, we will typically need to
find efficiently all arcs contained in some arc or all arcs
that might overlap some arc. Such queries can be answered by
computing the transitive closure $TA$ of the arc relation, but this is
likely to be an expensive proposition ($O(n^2)$ in the number of
nodes).  An alternative is to store the
two relations below.\footnote{
  Our approach has similarities with Allen's ``reference intervals''
\cite{Allen83}.
}
The relation {\it time} contains, for every
node $n$, the maximum time \ANTE\ of a timed node that precedes
$n$ and the minimum time \POST\  of a timed node that precedes
$n$.   If $n$ is itself timed, the {\em ante} and {\em post}
agree.\footnote{Some saving in space could be achieved by having a
separate relation for the timed nodes.}  It is a consequence of the
definitions in section \ref{ag-defs} that these times always
exist.  (Every node is bounded
by some pair of timed nodes.)  Note that {\em node} is a key for
the {\em time} relation, and we shall refer to the attributes \ANTE\ 
and \POST\ functionally, as $\ANTE(n)$ and $\POST(n)$.

The relation $TA'$ is defined by
\(
TA' = \{(m,n) | TA(m,n) \wedge \POST(m) > \ANTE(n)\}
\).
This means that the precedence relation $TC$
can be reconstructed by the query:
$$TC(m,n) :- \POST(m) < \ANTE(n) \vee TA'(m,n)$$

With indexes on \ANTE, \POST, and $(\mbox{source},\mbox{target})$,
this predicate can be efficiently computed.

The point of this decomposition is that we expect the relation $TA'$
to be relatively small.  For example, in the Switchboard database
\cite{Godfrey92}, the maximum size of $TC$ for any timeline is
approximately 1.9 million, while while the sizes of {\it time} and
$TA'$ are, for this timeline, $1,992$ and $10,585$
respectively.\footnote{
  This computation is based on the version of Switchboard data
  that is marked with time information at turn boundaries only.
  Given an $n$-word turn, the size of the transitive precedence
  relation is approximately $n^2/2$.
}
Throughout the whole database, the largest value of $TA'$ was
$15,286$.  Evidently the decomposed representation will
easily fit into main memory, while keeping $TC$ in main memory may
pose problems.

Finally, let us put together the ideas of the last two sections.  Consider
example (a) of the previous section.  The important point is that all
nodes are bounded by a \smtt{sentence} arc.   This suggests the
following technique:
\begin{itemize}
\item Repeatedly match\\
   $X$.\smtt{[type = parse, label = sentence].}$Y$
\item For each match, obtain $X'=\ANTE(X)$ and $Y'=\POST(Y)$
\item Restrict the arc relation to arcs bounded by $(X',Y')$
(use an index that supports range searches)
\item Perform the query on the restricted relation (main memory
evaluation should be possible)
\end{itemize}

\section{Conclusions}

Like semistructured data, annotation graphs have a natural
representation in terms of nodes and arcs.  A key feature
of annotation graphs is that the arcs are organized into a
quasi-linear flow in the horizontal direction.
As in the case of semistructured data, we seek a natural
query language for accessing and transforming this data.

This paper has described progress on a query language for
annotation graphs.  Path patterns and some abbreviatory devices
provide a convenient way to express a wide range of queries.
We exploit the quasi-linearity of annotation graphs by
partitioning the precedence relation, and we believe that
this will enable efficient temporal indexing of the graphs.

In ongoing work we are exploring hybrid structures and
languages which would permit both the vertical and horizontal
perspectives on semistructured data to co-exist.  On this view,
a horizontal path expression could be embedded inside a vertical
path expression, or vice versa.

\raggedright
\bibliographystyle{lrec2000}
\bibliography{general} 

\end{document}